%% file: NODE.tex
\documentclass{article}

\usepackage{arxiv}
\usepackage[utf8]{inputenc} 
\usepackage[T1]{fontenc}    
\usepackage[numbers,sort&compress]{natbib}      
\usepackage{url}            
\usepackage{booktabs}       
\usepackage{amsfonts}       
\usepackage{nicefrac}       
\usepackage{microtype}      
\usepackage{lipsum}		
\usepackage{graphicx}
\usepackage{natbib}
\usepackage{doi}
\usepackage{algorithm}
\usepackage{algorithmic}
\usepackage{enumitem}
\usepackage{amsmath}
\usepackage{amssymb}
\usepackage{multirow}
\usepackage{colortbl}
\usepackage{amsthm}
\usepackage{booktabs}
\usepackage{graphicx}
\usepackage{enumitem}
\usepackage{subcaption}
\usepackage{titletoc}
\input{math}

\usepackage{cleveref}
\crefname{equation}{Eq.}{Eqs.}
\Crefname{equation}{Equation}{Equations}
\crefname{figure}{Fig.}{Figs.}
\Crefname{figure}{Figure}{Figures}
\crefname{table}{Tab.}{Tabs.}
\Crefname{table}{Table}{Tables}
\crefname{section}{Sec.}{Secs.}
\Crefname{section}{Section}{Sections}
\crefname{proposition}{Prop.}{Props.}
\Crefname{proposition}{Proposition}{Propositions}
\crefname{appendix}{App.}{Apps.}
\Crefname{appendix}{Appendix}{Appendices}

\title{A Unified Control Framework for Gradient Inversion Defense in Federated Learning}

\author{Lianhai Ren \\
	Department of Mathematics\\
	National University of Singapore\\
	10 Lower Kent Ridge Road, Singapore 119076 \\
	\texttt{lianhairen@u.nus.edu} \\
	\And
	{Chi Zhang} \\
	Department of Mathematics\\
	National University of Singapore\\
	10 Lower Kent Ridge Road, Singapore 119076 \\
	\texttt{czhang24@nus.edu.sg} \\
}

\renewcommand{\shorttitle}{A Unified Control Framework for Gradient Inversion Defense in Federated Learning}

\hypersetup{
pdftitle={A Unified Control Framework for Gradient Inversion Defense in Federated Learning},
pdfsubject={q-bio.NC, q-bio.QM},
pdfauthor={Lianhai Ren, Chi Zhang},
pdfkeywords={First keyword, Second keyword, More},
}

\begin{document}
\maketitle

\begin{abstract}
	Gradient perturbations are widely recognized as an effective defense against gradient inversion attacks in federated learning, but they are equally known to impair optimization, creating a persistent tension between privacy and utility. In this paper, we show that existing gradient perturbation defenses, despite their different formulations, can be unified under a controlled dynamics framework, providing a cohesive understanding of their optimization behavior and enabling the derivation of universal convergence and safety conditions without relying on case-specific analyses. Leveraging this observation, we propose ``Near Orthogonal Gradient Projection (NODE)'', which injects perturbations only along directions orthogonal to the original gradient so that the defended gradient maintains a fixed cosine similarity with the true gradient. By preserving the first-order descent direction, NODE substantially reduces the optimization cost of strong perturbations while maintaining effective protection against gradient inversion attacks. We theoretically establish convergence guarantees under standard optimization assumptions and further show that, under mild conditions on the gradient distribution, NODE remains effective in preventing inversion attacks. Comprehensive experiments on multiple benchmark datasets, including CIFAR-10, ImageNet, and Chest X-ray, together with a variety of gradient inversion attacks, demonstrate that the proposed framework achieves an improved privacy--utility trade-off over existing defenses across a broad range of training settings.
\end{abstract}


\section{Introduction}
The exchange of gradients in federated learning inherently exposes the training process to privacy leakage, enabling adversaries to recover sensitive information from shared gradients. This vulnerability has been convincingly demonstrated by gradient inversion attacks~\cite{zhu2019deep,geiping2020inverting,zhang2023generative}, which can reconstruct training samples from transmitted gradients with remarkable fidelity. These findings fundamentally challenge the conventional assumption that federated learning preserves data privacy simply by keeping raw training data on local devices.

To mitigate this privacy risk, a wide range of defense mechanisms have been proposed, including gradient pruning~\cite{huang2021evaluating}, gradient modification~\cite{sun2021soteria}, gradient perturbation~\cite{abadi2016deep}, and differential privacy~\cite{dwork2008differential, dwork2014algorithmic}. Despite their diverse formulations, these methods share a common objective: preventing gradient inversion while preserving the utility of federated learning. Most existing approaches achieve this objective by deliberately modifying or restricting the information contained in transmitted gradients, seeking to strike an effective balance between privacy protection and optimization performance.

Yet, their analyses have largely remained fragmented, with each method evaluated in isolation. In this work, we demonstrate that these algorithms can be cast into a unified theoretical framework through the lens of controlled dynamics. Specifically, by modeling gradient modifications as instances of a generalized control function, we show that many existing defenses naturally fit within this overarching structure. This unification not only provides deeper insight into their underlying principles but also enables a systematic evaluation of their effectiveness, eliminating the need for ad hoc, method-specific analyses. We establish a comprehensive convergence and safety analysis for these algorithms, demonstrating how the design choices influence the convergence behavior of the original dynamics and assessing whether they offer theoretical guarantees for safeguarding against potential attacks.

Building upon this unified framework, we introduce the ``Near Orthogonal Gradient Projection'' (NODE) method, a defense strategy that offers a refined approach to injecting noise into gradients. Unlike conventional DPSGD approaches that introduce isotropic noise, NODE strategically injects noise solely in the subspace orthogonal to the true gradient. This design choice ensures that the first-order descent direction remains intact while maintaining a fixed cosine similarity between the true and the defended gradients. As a result, it ensures that model convergence remains unaffected, even as gradient magnitudes decrease. Furthermore, under mild assumptions on the gradient distribution, we theoretically establish that NODE remains highly resistant to gradient inversion attacks, effectively safeguarding sensitive training data without compromising convergence.

Extensive experiments on multiple datasets further demonstrate the effectiveness of NODE, showing that it maintains competitive test accuracy while substantially mitigating the success of diverse gradient inversion attacks, including GI~\cite{zhu2019deep}, IG~\cite{geiping2020inverting}, iDLG~\cite{zhao2020idlg}, CI-Net~\cite{zhang2023generative}, and DGGI~\cite{wu2025dggi}.

Our contributions can be summarized as follows:
\begin{itemize}
	\item Our work establishes a unified controlled dynamics framework that encapsulates a wide range of existing defense algorithms, revealing their underlying connections through a common theoretical lens. This perspective enables a systematic analysis of both convergence properties and privacy guarantees, offering a principled approach to evaluating different defense strategies in image and imaging applications.
	\item We introduce NODE, a new defense mechanism that employs orthogonal noise injection to maintain a near-orthogonal alignment with the gradient, thereby preserving the core descent direction while fortifying image models against gradient inversion attacks.
	\item We conduct a theoretical analysis of NODE under standard convexity conditions and fixed learning rates, proving that it ensures stable convergence while delivering robust defense against inversion algorithms. This theoretical foundation is supported by empirical results on standard image-processing and medical imaging benchmarks, demonstrating its effectiveness in practical imaging scenarios.
\end{itemize}

\section{Literature Review}
\subsection{Gradient Inversion Attacks}
Federated Learning (FL)~\cite{konevcny2016federated, mcmahan2017communication} has become a widely adopted paradigm for distributed model training, especially for modern over-parameterized large models~\cite{zhang2021distributed}, by enabling collaborative optimization while keeping raw data on local devices. However, this data-localization mechanism alone does not ensure privacy, as the gradients exchanged between clients and servers may contain rich information about the underlying training samples. Gradient inversion attacks~\cite{zhu2019deep} demonstrate this vulnerability by exploiting shared gradients to reconstruct private data, revealing that model updates can serve as an effective medium for unintended information leakage. Early approaches established the feasibility of recovering training samples by jointly optimizing dummy inputs and labels to match the observed gradients~\cite{zhu2019deep,zhao2020idlg}. Subsequent methods improved reconstruction fidelity by incorporating prior information, including natural image statistics~\cite{geiping2020inverting,yin2021see} and architectural characteristics of target models. In parallel, analytical studies have investigated the theoretical conditions under which gradients can be explicitly inverted, providing deeper insights into the underlying leakage mechanisms~\cite{geng2021towards,boenisch2023curious}. More recently, generative models, such as GANs and diffusion models~\cite{hatamizadeh2022gradvit,wu2025dggi,sotthiwat2024generative}, have further advanced reconstruction quality and scalability by introducing powerful data priors~\cite{jeon2021gradient,zhang2023generative,chen2024gifd}. Beyond the original single-sample reconstruction setting, gradient inversion attacks have expanded to more challenging scenarios, including large-batch recovery~\cite{balunovic2022lamp}, privacy leakage accumulated over multiple communication rounds~\cite{fowl2022decepticons}, and cross-domain applications such as language models, where gradients may expose sensitive textual information~\cite{gupta2022recovering,deng2021tag}.

\subsection{Defenses Against Inversion}
Defensive strategies fall into several categories. Cryptographic solutions such as multi-party computation (MPC) and homomorphic encryption (HE) offer strong theoretical protection by avoiding clear-text gradient exposure, but they incur substantial communication and computation overhead~\cite{yao1982protocols, rivest1978data, zhao2019secure, fereidooni2021safelearn, ma2022privacy}. Secure aggregation protocols provide a practical middle ground by enabling aggregation without revealing individual gradients~\cite{bonawitz2017practical, bell2020secure,zhang2022augmented,sotthiwat2021partially}. Differential privacy (DP) and DPSGD introduce calibrated noise to obscure individual contributions and are more scalable for large FL deployments~\cite{dwork2008differential, dwork2014algorithmic, abadi2016deep, mironov2017renyi, wei2020federated,teo2025privacy}, though recent work has explored adaptive noise mechanisms to improve the utility-privacy tradeoff~\cite{noble2022adaptive, kurakin2022toward}. Complementary approaches include pruning or sparsification to remove informative gradient components~\cite{huang2021evaluating, zhu2021gradient}, gradient compression techniques that reduce information leakage~\cite{wang2021sapag, scheliga2022precode}, and adaptive masking or representation transformations (e.g., Soteria) that aim to hide reconstructible features while retaining utility~\cite{sun2021soteria, huang2021instahide}. Model perturbation methods such as mixup-based defenses~\cite{shen2022mixup} and input preprocessing techniques~\cite{li2022auditfl} have also been proposed to complicate reconstruction. Despite diverse mechanisms, these defenses share the core goal of transforming gradient information to impede inversion while preserving model training for image or video tasks, though many face challenges in defending against adaptive attacks~\cite{geiping2021breaching, balancing2022tramer}.

\subsection{Control Perspectives for Neural Network Training}
Viewing optimization and deep networks through the lens of dynamical systems and control offers a unifying perspective. Residual networks and continuous-depth models can be interpreted as discrete-time dynamical systems, with parameters or gradient modifications acting as control inputs that steer the evolution of model states~\cite{weinan2017proposal, li2017maximum, he2016deep}. Control and feedback theory have informed architecture design, stability analysis, and optimization methods in deep learning~\cite{franklin2002feedback,doyle2013feedback, benning2019deep, tabuada2020universal,zhang2024parameter,zhang2024optimal,zhang2025weight}. Building on this viewpoint, we cast a broad class of gradient-modification defenses as instances of controlled dynamics, which allows systematic comparison of convergence and privacy properties and removes the need for fragmented, method-specific proofs.

\section{A Unified Framework for Defense Against Gradient Inversion Attacks}
In this section, we first revisit the gradient inversion attack in FL, emphasizing the privacy challenges inherent in the FL paradigm. We then show that many existing defense methods, such as DP, pruning, and Soteria, can be interpreted as functions acting on the gradients (and, optionally, on the input data). This observation motivates our unified framework, which treats the gradient defense operation as a controller within the training dynamics.

\subsection{Gradient Inversion Attack in Federated Learning}
In conventional FL systems, local training data and associated labels are generally not accessible to external entities, such as the curious server or third parties. Instead, these entities may only gain access to the uploaded local gradients. The gradient inversion attack~\cite{zhu2019deep} aims to reconstruct the original data from these gradients by leveraging the relationship between the model's weights, the loss function, and the gradients. 

Mathematically, consider a model with parameters $w$ and a loss function $L(w, x)$, where $x$ represents the input data. In the case of the gradient inversion attack, the adversary seeks to recover $x$ by solving an inverse problem. Given the gradient $g$, the attacker aims to approximate the original input $x$, which can be formulated as an optimization problem:

\[
\hat{x} = \arg \min_{x^\prime} \| \nabla_w L(w, x^{\prime}) - g \|^2.
\]

Advanced techniques, such as regularization~\cite{geiping2020inverting} or priors~\cite{jeon2021gradient}, are often employed by attackers to improve the quality of the reconstructed data and reduce the ambiguity inherent in the inversion process. Recent work~\cite{zhang2023generative} has demonstrated the feasibility of such attacks, even with a relatively large batch size, highlighting the vulnerability of current FL training paradigms to privacy breaches.

\subsection{A Unified Framework for Defense Through Controlled Dynamics}
To reduce the risk of gradient inversion attacks, various defense mechanisms have been proposed to strengthen communication security. Common approaches include DPSGD~\cite{abadi2016deep}, which leverages differential privacy to protect data, pruning techniques~\cite{huang2021evaluating} that selectively eliminate a substantial portion of gradients, and Soteria~\cite{sun2021soteria}, which seeks to generate a perturbed data representation to safeguard sensitive information.

Despite their apparent differences, we demonstrate that these methods can be cast into a unified framework through the lens of control theory. Formally, let $x_t\in\R^n,w_t\in\R^m$ denote the input and the weight at the $t$-th iteration, $g_t$ denote the corresponding gradient, and $C(g,x):\R^n\times\R^m\to\R^m$ denote the control function applied to the gradient by the defense method. We can express a general defense operation as controlled dynamics in the training:
\begin{equation}
	\frac{dw_t}{dt}=C(g_t,x_t):=g'_t.
	\label{eq: modified dynamics}
\end{equation}

Within the framework of control dynamics, many existing defense methods can be viewed as specific instances of selecting a particular control function $C$:
\begin{itemize} 
	\item \textbf{DPSGD} implements the transformation as
	\begin{equation}
		C_{\text{DP}}(g,x)=g+\xi,
	\end{equation}
	where $\xi\in\R^m$ represents noise sampled from a predefined distribution, typically Gaussian or Poisson. For simplicity, we assume that DPSGD utilizes a Gaussian distribution throughout this paper.
	\item \textbf{Pruning} can be interpreted as a function that zeros out or attenuates certain components of the gradient based on their magnitude or importance:
	\begin{equation}
		C_{\text{prune}}(g,x;k)=g\odot\mathbb{I}\{|g_i|\geq \text{Top}_k(|g_i|)\},
	\end{equation}
	where $\odot$ denotes element-wise multiplication, $\mathbb{I}$ represents the indicator function, $\text{Top}_k(|g_i|)$ returns the $k$-th largest absolute value  among the gradient components.
	\item \textbf{Soteria} utilizes the gradient norm of the input to the output features (specifically, the input to the last fully connected (FC) layer) to generate a mask, which is then applied to the original gradient of the last FC layer. This can be expressed as:
	\begin{equation}
		C_{\text{soteria}}(g,x)=g\odot F_{\text{soteria}}(g,x),
	\end{equation}
	where $F_{\text{soteria}}(g,x)$\footnote{For detailed definitions on $F_{\text{soteria}}$, please refer to Section 4 of \cite{sun2021soteria}.} is a function that generates masks. 
\end{itemize}


This unified control perspective implies that, despite the diversity of defense mechanisms, these approaches can be viewed as applying controlled modifications to the gradient information. Furthermore, this perspective enables the evaluation of various defense strategies within a shared theoretical framework, thereby obviating the necessity for individual, method-specific analysis. By conceptualizing different defense methods as distinct control strategies, the framework facilitates generalization across various techniques, thus providing a foundation for the development of new defense methods, as discussed in Section~\ref{sec:4}.
\subsection{General Conditions for Defense Algorithms}
In general, an effective defense algorithm must achieve two key objectives: it should preserve model performance within the training dynamics while safeguarding data against gradient inversion attacks. With the unified framework in place, we are now able to pose a key question: \textit{What properties must the control function $C$ possess to ensure these two conditions are met?}


\subsubsection{Convergence Condition} 
We begin by examining how the selection of the defense operation $C(g,x)$ influences the convergence properties of the training algorithm. In particular, the final performance of the modified training dynamics should closely approximate that of the original dynamics. Formally, if $w_\infty$ and $w_\infty^o$ denote the final weights obtained from the modified and original dynamics respectively, we require:
\begin{equation}
	\E[L(w_\infty)-L(w_\infty^o)]\leq \ell_0,
\end{equation}
where $L:\R^m\to\R$ is the loss function of weight, i.e., the average loss of the whole dataset, and the expectation is taken over the initial weights. A smaller $\ell_0$ indicates that the defense method maintains training performance effectively.

For example, we can analyze the value of $\ell_0$ under $c-$strongly convex loss function $L(w)$ with $l-$Lipschitz continuous gradient, i.e.:
\begin{align}
	& L(w')-L(w) \geq \nabla L(w)^\top(w'-w)+\frac{c}{2}\|w-w'\|^2_2,
	\label{eq: strongly convex}\\
	&\|\nabla L(w)-\nabla L(w')\|_2\leq l\|w-w'\|_2.
	\label{eq: lipschitz gradient}
\end{align}
With simple transformations, we can derive from Eq \eqref{eq: strongly convex}:
\begin{equation}
	L^*\geq L(w)-\frac{c}{2}\|\nabla L(w)\|.
	\label{eq: strongly convex col}
\end{equation}
Additionally, Eq \eqref{eq: lipschitz gradient} implies the following bound:
\begin{equation}
	\begin{aligned}
		L(w')
		\leq L(w)+\nabla L(w)^\top(w'-w)+\frac{l}{2}\|w'-w\|_2^2.
	\end{aligned}
	\label{eq: lipschitz col}
\end{equation}

For simplicity, here we focus on the GD algorithm. Since defense methods often add random noise on the gradient or selectively zero-out certain components, we need some general first and second moments limit on the defended gradient $g'$ in the following Assumption.
\begin{assumption}(Assumption 4.3 in \cite{bottou2018optimization}) 
	\label{assumption: grad angle}
	Assume the defended gradient \( g' \) satisfies the following first and second moments constraints, $\exists\ \mu>0,\mu_G>0,M\geq 0,M_V\geq 0$, such that:
	\begin{align}
		& \nabla L(w)^\top \mathbb{E}[g'] \geq \mu\|\nabla L(w)\|_2^2, 
		\label{eq: moment 1} \\
		& \| \mathbb{E}[g']\|_2 \leq \mu_G\|\nabla L(w)\|_2, 
		\label{eq: moment 2} \\
		& \text{Var}(g') \leq M + M_V\|\nabla L(w)\|_2^2. 
		\label{eq: moment 3}
	\end{align}
\end{assumption}

Using the above assumptions in Eqs \eqref{eq: strongly convex col}–\eqref{eq: moment 3}, we can then have the following theorem for the convergence.
\begin{theorem}
	\label{thm: convergence}
	Assume the loss function is $c$-strongly convex with an $l$-Lipschitz continuous gradient, and the defended gradient $g'$ satisfies the moment constraints \eqref{eq: moment 1}-\eqref{eq: moment 3} uniformly in $w$. If the learning rate is fixed within the range $0<\eta\leq\frac{\mu}{lM_G}$, where $M_G=M_V+\mu_G^2$, then the following convergence guarantee holds:
	\begin{equation}
		\label{eq: converge}
		\begin{aligned}
			\E[L(w^k)-L^*]&\leq\frac{\eta lM}{2c\mu}+(1-\eta c\mu)^k\left(L(w^0)-L^*-\frac{\eta lM}{2c\mu}\right)\\
			&\stackrel{k\to\infty}\longrightarrow \frac{\eta lM}{2c\mu},
		\end{aligned}
	\end{equation}
	where $w^k$ denotes the weight after $k$ steps and the expectation is conditioned on initial weight $w^0$.
\end{theorem}
We can further extend the result to the non-convex case, similar to GD:
\begin{theorem}
	\label{thm: convergence nonconvex}
	Assume the same conditions in Theorem~\ref{thm: convergence} without the convexity condition for the loss function, then the following convergence guarantee holds:
	\begin{equation}
		\label{eq: converge}
		\begin{aligned}
			\E\left[\frac{1}{K}\sum_{k=1}^K\|\nabla L(w^k)\|_2^2\right]&\leq\frac{\eta lM}{2c\mu}+\frac{(L(w^0)-L^*)}{Kc\mu}\\
			&\stackrel{k\to\infty}\longrightarrow \frac{\eta lM}{2c\mu},
		\end{aligned}
	\end{equation}
\end{theorem}
\begin{remark}
	The above two theorems suggest that if the upper bound on the variance of the protected gradient $g'$ includes a constant factor $M$ in Eq \eqref{eq: moment 3}, then we have a constant factor $\frac{\eta lM}{2c\mu}$ in the infinite time. Moreover, when the true gradient is small, this factor will dominate the defended gradient, resulting in a nonzero gap between the final loss and the optimal loss.
\end{remark}
\begin{remark}
	\label{remark: perpendicular}
	Note that in Assumption~\ref{assumption: grad angle}, it requires the defended gradient $g'$ and the true gradient to have an angle that is lower and away from $90^\circ$. We have proven that for convex loss, using a perpendicular defended gradient will not decrease the value.\footnote{In~\cite{zhang2025censor}, authors propose perpendicular defended gradient. We notice an error in their defense function that causes the defended gradient to exhibit positive cosine similarity with the true gradient. After correcting this issue, we find that their training procedure may no longer succeed.}
\end{remark}

\subsubsection{Safety Condition} 
We now turn our attention to the safety condition that an effective defense algorithm must satisfy. In particular, the control function in Eq~\eqref{eq: modified dynamics} must sufficiently obfuscate sensitive gradient information to protect against inversion attacks. This means that $C(g,x)$ should distort the gradient in a controlled manner so that an adversary cannot easily recover the underlying data $x$. Formally, let $G_w^{-1}$ denote the inverse mapping of the input-gradient function under weight $w$\footnote{Here we assume the input-gradient function $G_w$ is surjective.}. Then, for all $x$ in the data domain $D\subseteq\R^n$ (e.g., the private dataset or a bounded set), we require:
\begin{equation}
	\min_{x'\in G_w^{-1}(C'(g,x))}\|x'-x\|_2\geq \ell_1,\forall x\in D,\forall w\in W\backslash W_\epsilon.
\end{equation}
where $W\subseteq\R^m$ is the parameter set and $W_\epsilon=\{w|\|\nabla L(w,x)\|_2\leq \epsilon,\forall x\in D\}$ is the set near the saddle point where the gradient is small such that it is near impossible to attack in $W_\epsilon$~\cite{geiping2020inverting}.  A larger $\ell_1$ implies stronger obfuscation and, hence, better defense performance.

A comprehensive analysis of the $\ell_1$ is generally non-trivial and often contingent on the specific formulation of the gradient inversion attack. But we can show that it can be bound by the Lipschitz constant of the input-gradient map:
\begin{theorem}
	\label{thm: in-grad Lipschitz}
	If the input-gradient function $G_w:x\to\nabla L(w,x)$ is $l_g$ Lipschitz continuous for all $w\in W$, then we have:
	\begin{equation}
		\ell_1\geq\frac{1}{l_g}\min_{x\in D}\|C(g,x)-g\|_2.
	\end{equation}
\end{theorem}
In conclusion, a gradient defense method that achieves a small $\ell_0$  (ensuring minimal degradation in training performance) and a large $\ell_1$ (ensuring robust privacy protection) is considered effective. This perspective unifies the analysis of existing defense methods and provides clear guidelines for designing new, more robust defenses in FL systems.
\subsection{Can Existing Defense Methods Satisfy These Conditions?}
\label{sec: theory DPSGD}
By establishing a unified framework along with the corresponding convergence and safety conditions, we are now positioned to assess whether existing defense algorithms adhere to these criteria. 

We begin by analyzing the convergence and safety conditions of the DPSGD algorithm in the following Theorems.

\begin{theorem}[$\ell_0$ Bound for DPSGD]
	\label{thm: DP converge}
	Under the assumptions \eqref{eq: strongly convex col}–\eqref{eq: moment 3}, for DPSGD with noise scale $\sigma$ and $l_2$-norm clipping threshold $a$, we have:
	\begin{equation}
		\ell_{0,\text{DPSGD}}\leq \frac{\eta la^2\sigma^2}{2c}.
	\end{equation}
\end{theorem}
\begin{remark}
	Theorem \ref{thm: DP converge} indicates that, in general, we can only obtain a constant factor when comparing the performance of DPSGD relative to the original dynamics. Convergence to the original dynamics is therefore not possible with $\sigma > 0$. This gap arises from the added noise in the gradients. Moreover, a larger noise scale $\sigma$ and $l_2$-norm clipping threshold lead to worse model performance, which aligns with both intuitive expectations and previous experimental findings.
\end{remark}

In terms of the safety condition, we have $\min\|C(g,x)\|_2=\min\|\xi\|_2=0$  for DPSGD. However, we can obtain the lower bound of $\ell_1$ with $1-2\delta$ confidence:
\begin{theorem}[$\ell_1$ Bound for DPSGD]
	\label{thm:ell_1 for DPSGD}
	Under the assumptions above, for DPSGD with noise scale $\sigma$ and $l_2$-norm clipping threshold $a$, with $1-2\delta$ probability, we have:
	\begin{equation}
		\ell_{1,\text{DPSGD}}\geq\frac{a\sigma (\sqrt{m}-a_0\sqrt{\log \delta^{-1}})}{l_g}.
	\end{equation}
	Here $a_0$ is an absolute constant.
\end{theorem}

In summary, Theorem \ref{thm:ell_1 for DPSGD} establishes that DPSGD can be considered a safe defense method when the noise scale \( \sigma \) is relatively large. However, this comes at the cost of violating the convergence condition: the controlled dynamics may exhibit a performance gap compared to the original dynamics, as highlighted in Theorem \ref{thm: DP converge}. In practice, a careful choice of \( \sigma \) is necessary to balance the trade-off between convergence and safety.


Similarly, for the pruning method, we can establish the following two theorems.
\begin{theorem}[$\ell_0$ for Pruning]
	\label{thm:ell_0 for Pruning}
	Under the assumptions \eqref{eq: strongly convex col}, \eqref{eq: lipschitz col}, for the pruning method with pruning rate $p < 1$, we have:
	\begin{equation}
		\ell_{0,\text{Pruning}}=0.
	\end{equation}
\end{theorem}

\begin{remark}
	The above theorem suggests that, under appropriate assumptions, the pruning method can achieve the same final loss as the original optimization dynamics, thereby satisfying the convergence condition. This outcome is expected, as even in the extreme scenario where only a single element is retained, the pruning reduces to the coordinate descent~\cite{nesterov2018lectures}, which is well-established to converge for convex functions~\cite{wright2015coordinate,tseng2001convergence}.
\end{remark}
To analyze the $\ell_1$ for the pruning method, we assume that
\begin{align}
	g \sim N(0,\sigma_g^2 I_m).\label{assumption: gaussion grad}
\end{align}
To continue, we consider an operation $a$-truncation on a random variable $X$, by which we obtain $\hat{X}=0$ if $|X|> a$ and $\hat{X}=X$ otherwise. Then we have the Theorem.
\begin{theorem}[$\ell_1$ Bound for Threshold Pruning]
	\label{thm:ell_1 for Pruning}
	Under the assumptions \eqref{eq: strongly convex col}, \eqref{eq: lipschitz col}, \eqref{assumption: gaussion grad}, for the threshold pruning method with pruning threshold $a$, with $1-2\delta$ probability, we have:
	\begin{equation}
		\ell_{1,\text{Threshold Pruning}}\geq \frac{\sigma_g \sqrt{V(a)} (\sqrt{m}-a_1\sqrt{\log \delta^{-1}})}{l_g}.
	\end{equation}
	Here $V(a)$ is the variance of $a$-truncated standard normal random variable and $a_1$ is an absolute constant.
\end{theorem}
\begin{remark}
	It is generally non-trivial to directly analyze the $\ell_1$ for the pruning method, but with the independence of $a$-truncated random variables, it is easier to analyze the pruning method with a threshold. As $m\to \infty$, the frequency of pruning ($|X_i|\leq a$) almost surely converges to the pruning rate $p$ with the speed of $O(m^{-\frac{1}{2}})$, by which we can infer $a$ from the expected rate $p$: $a=\Phi^{-1}(\frac{1+p}{2})$. Here $\Phi$ is the CDF of standard normal distribution. As $p \to 0$, the $\ell_1$ norm can become relatively small, potentially increasing the risk of local data leakage to attackers.
\end{remark}
In summary, despite satisfying the convergence condition in Theorem~\ref{thm:ell_0 for Pruning}, the pruning method fails to guarantee the safety requirement in Theorem~\ref{thm:ell_1 for Pruning}, as it does not establish a sufficiently large $\ell_1$ loss.

\section{Near Orthogonal Gradient Projection with Convergence and Safety Guarantees}
\label{sec:4}
The violation of either the convergence or safety condition for previous algorithms highlights the necessity of a defense mechanism that simultaneously satisfies both. As such, we introduce our proposed method—Near Orthogonal graDient projEction (NODE)—and provide theoretical analysis to guarantee that NODE both preserves training performance and maintains robust defense in this section.


\subsection{Key Observations on DPSGD}
Before introducing our new algorithm, we first present two key observations that explain why DPSGD fails to meet the convergence condition.

\textbf{Observation 1: DP Noise Scale Trade-off.}
In traditional DPSGD methods, increasing the noise scale improves defense performance by better obscuring sensitive information. However, a large noise scale also distorts the gradient direction, which can lead to significant performance degradation. This trade-off becomes especially critical when a strong defense is required, yet maintaining high training performance is also desired.

\begin{figure}[t]
	\centering
	\includegraphics[width=.6\linewidth]{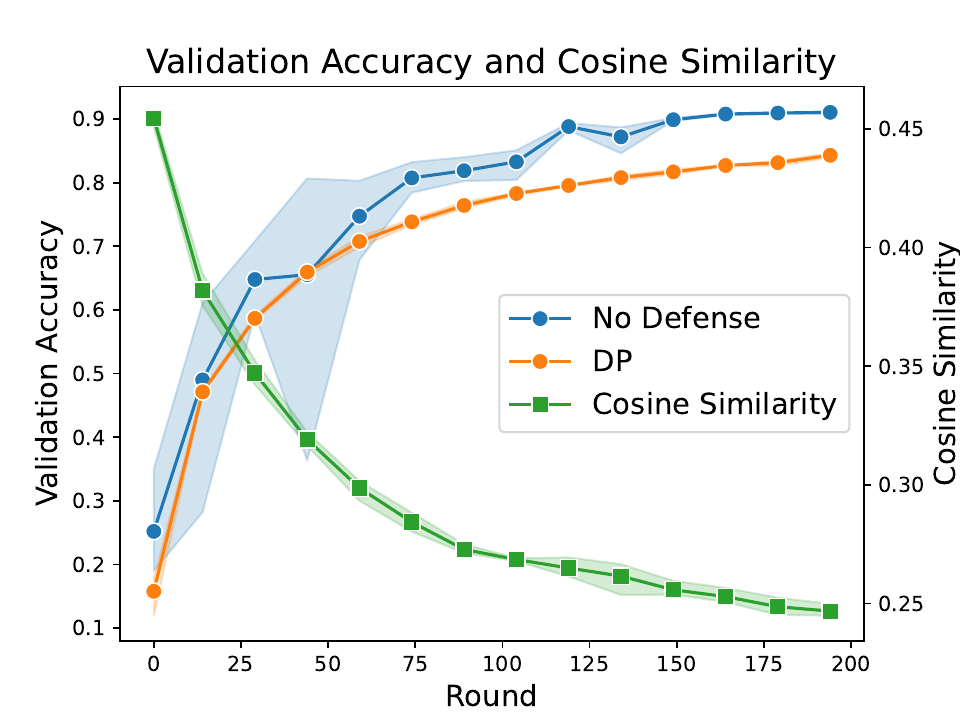}
	\caption{During training, the DPSGD method will deviate gradient more since the true gradient becomes small (green curve), resulting in a performance drop when compared to the original dynamics. \textbf{Left Axis:} Accuracy; \textbf{Right Axis:} Cosine Similarity.}
	\label{fig:train_angle}
\end{figure}

\textbf{Observation 2: Late-Stage Training Instability.}
Empirically, DPSGD methods behave similarly to the original dynamics during the early stages of training when the true gradients are relatively large. As training progresses, the true gradients become smaller, and the fixed noise level can dominate the signal. This imbalance causes the cosine similarity between the defended gradient and the original gradient to decrease over time (the green curve in Figure~\ref{fig:train_angle})—that is, the angle between them gradually increases—resulting in instability and degraded model performance in later stages.

\begin{algorithm}[t] \caption{Near Orthogonal Gradient Projection (NODE)} 
	\label{alg:NODE} 
	\begin{algorithmic}[1] 
		\REQUIRE Client gradient $g_t\in\R^d$, desired angle $\theta$ \ENSURE defended gradient $g_t'$
		\STATE Compute the norm of the gradient: $G_t\leftarrow\|g_t\|_2$
		\STATE Sample a noise vector $\xi_t\sim\mathcal{N}(0,I_d)$
		\STATE Project $\xi_t$ onto the hyperplane orthogonal to $g_t$: $n_t'\leftarrow \xi_t-\frac{\langle \xi_t,g_t\rangle}{G^2}g_t$
		\STATE Set the noise magnitude: $r_t\leftarrow\frac{G_t}{\|n'_t\|_2}\tan\theta$
		\STATE Compute the defended gradient: $g_t'\leftarrow g_t+r_tn_t'$
		\RETURN $g_t'$
	\end{algorithmic} 
\end{algorithm}
\subsection{Near Orthogonal Gradient Projection (NODE)}
The above observations suggest that, during the later stages of training when the gradients are inherently small, the same noise scale used earlier overestimates the necessary perturbation. Consequently, the noise should be injected in a controlled manner—preserving the descent direction while still ensuring robust defense. To accomplish this, there are two cornerstones: first, the injected noise must remain orthogonal to the true gradient direction, ensuring minimal interference with the descent path; second, the defended gradient should maintain a fixed angular relationship with the true gradient, preventing instability in later stages. 

Specifically, we introduce a ``Near Orthogonal Gradient Projection'' (NODE) approach that injects noise strictly in the orthogonal direction of the gradient and enforces a controlled clipping mechanism to ensure the modified gradients remain nearly orthogonal to the original ones. For each iteration $t$, the defended gradient $g_t'$ is constructed by adding a noise vector $n_t$ that is strictly orthogonal to the true gradient $g_t$. This ensures that, at least in a first-order approximation, the descent direction remains unaffected. Moreover, by scaling the noise proportionally with $\|g_t\|$, NODE maintains a constant cosine similarity (e.g., $0.2$) between $g_t$ and $g_t'$. Mathematically, we enforce:
\begin{align*}
	&\quad \ \ C(g_t)=n_t, \\
	& \mbox{s.t. } \langle g_t,n_t\rangle=0, \cos\theta=\frac{\langle g_t,g_t'\rangle}{\|g_t\|_2\|g_t'\|_2}=\alpha.
\end{align*}
where $n_t$ is uniformly distributed in the accepted area (which is the intersection of a hyperplane and a hyper-spherical surface), $\alpha$  (or equivalently, the fixed angle $\theta$) is a predetermined constant that balances training performance and defense robustness. 

Algorithm \ref{alg:NODE} outlines the NODE procedure for a given client gradient. In this algorithm, the noise $n_t$ is generated by first sampling from a standard Gaussian distribution, then projecting onto the subspace orthogonal to $g_t$, and finally rescaling to ensure that its magnitude is proportional to $\|g_t\|_2$ (specifically, $\|n_t\|_2=\tan\theta\|g_t\|_2$ ). By adding $n_t$ to $g_t$, NODE preserves the descent direction while ensuring that the cosine similarity between $g_t$ and $g_t'$ remains fixed. This controlled noise injection is designed to thwart gradient inversion attacks without significantly impacting training performance.

\subsection{Theoretical Analysis}
\label{sec: theory NODE}
We now analyze the convergence properties of NODE based on both Theorem \ref{thm: convergence} and the dynamical view, and establish conditions under which it maintains both training performance and defense effectiveness.
\subsubsection{Convergence Condition}
We start by analyzing the convergence properties of NODE. In particular, we can use Theorem \ref{thm: convergence} to get the convergence result for NODE under strongly convex and first-order Lipschitz continuous loss:
\begin{theorem}[$\ell_0$ Bound for NODE]
	\label{thm:ell_0 for NODE}
	Under assumptions in Theorem \ref{thm: convergence}, NODE has $\ell_{0,\text{NODE}}=0$.
\end{theorem}
\begin{remark}
	Similar to the pruning method, the NODE method can achieve
	the same final loss as the original dynamics,
	thereby satisfying the convergence condition. 
\end{remark}

Moreover, we show it is possible to remove the assumptions ~\ref{eq: moment 1}-\ref{eq: moment 3} and get the same convergence result for NODE in continuous training dynamics with the Lyapunov function~\cite{lyapunov1992general} in the following theorem.
\begin{theorem}
	\label{thm: lyapunov}
	$(L(w)-L^*)^2$ is the Lyapunov function for the NODE method, so it will asymptotically reach the minimum $0$, i.e. the NODE method will asymptotically converge to the minimum.
\end{theorem}


\subsubsection{Safety Condition} 
Since the noise scale is proportional to the gradient scale, which decreases during training, we need to establish some mild additional conditions under which NODE maintains the defense effectiveness.
\begin{assumption}
	\label{assumption:gradient norm}
	Assuming the norm of gradient of each data $x$ in the dataset $D$ is bounded by the norm of the average gradient of the dataset, i.e. $\exists b>0,$ such that $\forall x\in D$ we have:
	\begin{equation}
		\|\nabla L(w;x)\|_2\leq b \|\nabla L(w)\|_2.
	\end{equation}
\end{assumption}
This assumption limits the gradient scale in the final training stage, where the overall gradient is small. With this assumption, we can prove that the NODE has the same DP budget as DPSGD.
\begin{theorem}
	\label{thm:defense}
	Under Assumption \ref{assumption:gradient norm}, NODE with angle $\theta$ has the same differential privacy budget as DPSGD with $\sigma=\frac{\tan\theta}{b}$.
\end{theorem}

\begin{remark}
	Theorem \ref{thm:defense} demonstrates that the NODE method theoretically exhibits a similar defense ability as the DPSGD method.  As a result, it is feasible to select an appropriate value for $\theta$ to ensure that $\ell_1$ remains sufficiently large, thereby satisfying the safety condition.
\end{remark}

In summary, our analysis indicates that the NODE method can reshape gradient information to safeguard privacy without crippling the learning performance with assumptions \eqref{eq: strongly convex col}–\eqref{eq: lipschitz col}, therefore satisfying the convergence and safety conditions simultaneously.

\section{Experiment}
\label{sec:5}
This section evaluates our method on multiple datasets, measuring its robustness against gradient inversion attacks. 

\subsection{Experiment Setup}
Our experimental setup follows common practices from prior gradient-inversion work~\cite{zhu2019deep,zhang2023generative}. We use ResNet-18 models with sigmoid activations, which have been shown to increase susceptibility to inversion attacks. We compare the following defense methods: (1) No Defense; (2) DP-SGD~\cite{abadi2016deep} with varying noise scales; (3) Pruning~\cite{huang2021evaluating} at different sparsity levels; (4) Soteria~\cite{sun2021soteria}; (5) Outpost~\cite{wang_more_nodate}; (6) Dual Gradient Pruning (DGP)~\cite{xue_revisiting_2024}; and (7) Gradient Dropout~\cite{sotthiwat2026safeguarding}; (8) our proposed NODE. To evaluate defense effectiveness, we consider five representative gradient-inversion attacks: Gradient Inversion (GI)~\cite{zhu2019deep}, Improved Gradient Inversion (IG)~\cite{geiping2020inverting}, iDLG~\cite{zhao2020idlg}, CI-Net~\cite{zhang2023generative}, and DGGI~\cite{wu2025dggi}. We report both model training performance and the degree of privacy leakage, measured using SSIM and FSIM. Detailed setup is provided in Appendix B.

\subsection{CIFAR-10 Results} 
We initiate our evaluation of defense performance on the CIFAR-10 dataset~\cite{krizhevsky2009learning}, presenting the results in Table~\ref{tab:cifar10_results}. The baseline model achieves a high test accuracy of 91.21\%. However, it exhibits high SSIM values under inversion attacks, indicating that the recovered images remain too similar to the original data. In contrast, defense methods such as pruning (p=$0.9$) and DP ($\sigma$=4e-3) achieve lower SSIM values, hence offering improved privacy protection. Yet, this often comes at the cost of reduced test accuracy when pruning a large portion of weights or increasing the noise level in DPSGD.  Notably, NODE with a cosine similarity of $0.2$ maintains competitive test accuracy ($90.30$\%) while drastically reducing the SSIM values (e.g., $0.0256$ under GI), demonstrating its effectiveness in obfuscating gradients. Further setting $\alpha=0.1$ decreases the SSIM values while resulting in only a minor $0.99\%$ drop in accuracy. Overall,  NODE strikes a favorable balance between preserving  performance and enhancing safety.

\begin{table*}[htbp]
	\centering
	\caption{Performance on CIFAR-10: Final test accuracy and average SSIM under four gradient inversion attacks (GI, IG, iDLG, CI-Net and DGGI) across various defense methods. The best performance is in bold, and the second-best is underlined.}
	\resizebox{\linewidth}{!}{
		\begin{tabular}{l|c|ccccc|c}
			\toprule
			\multirow{2}{*}{Defense Methods} & \multicolumn{1}{c|}{Performance} & \multicolumn{5}{c}{Attack (SSIM)} \\
			\cline{2-8}    \multicolumn{1}{c|}{} & Test Acc $\uparrow$  & GI $\downarrow$    & IG $\downarrow$   & iDLG $\downarrow$  & CI-Net $\downarrow$ & DGGI $\downarrow$ & Mean \\
			\midrule
			None  & \textbf{91.21}$_{\pm0.17}$ & $ 0.3751_{\pm 0.0929}$ & $ 0.4701_{\pm 0.0948}$ &  0.3974$_{\pm 0.0755}$ &  0.9866$_{\pm 0.0081}$ &  0.5017$_{\pm 0.1084}$ &0.5462 \\
			Pruning (p=0.9) &  89.66$_{\pm 0.17}$ &  0.0317$_{\pm 0.0103}$ &  0.0367$_{\pm 0.0208}$ &  0.0222$_{\pm 0.0083}$ &  0.8513$_{\pm 0.0814}$ & 0.1577$_{\pm 0.1040}$ &0.2199 \\
			DPSGD ($\sigma$=1e-3) &  87.09$_{\pm 0.28}$ &  0.0958$_{\pm 0.0484}$ &  0.4211$_{\pm 0.1235}$ &  0.3492$_{\pm 0.0752}$ &  0.9437$_{\pm 0.0196}$ & 0.4655$_{\pm 0.1038}$ &0.4551 \\
			DPSGD ($\sigma$=4e-3) & 85.06$_{\pm 0.26}$  & 0.1284$_{\pm 0.0308}$ & 0.0462$_{\pm 0.0249}$ & 0.1446$_{\pm 0.0392}$ & 0.2933$_{\pm 0.0989}$ & 0.2576$_{\pm 0.1224}$ &0.1740 \\
			Soteria &   89.20$_{\pm 0.15}$    &  0.1143$_{\pm 0.0506}$ &  0.1886$_{\pm 0.0225}$ &  0.2651$_{\pm 0.0064}$ &  0.4122$_{\pm 0.1203}$ & 0.2675$_{\pm 0.0817}$ &0.2495 \\
			Outpost & 61.68$_{\pm 1.35}$ & \textbf{0.0241}$_{\pm 0.0089}$ & 0.0389$_{\pm 0.0298}$ & 0.0311$_{\pm 0.0130}$ & 0.9402$_{\pm 0.0358}$ & 0.1657$_{\pm 0.1149}$ &0.2400 \\
			DGP  & 89.50$_{\pm 0.12}$ & \underline{0.0249}$_{\pm 0.0051}$ & \textbf{0.0255}$_{\pm 0.0073}$ & \textbf{0.0160}$_{\pm 0.0073}$ & 0.8080$_{\pm 0.1103}$ & \textbf{0.1515}$_{\pm 0.1074}$ &0.2052 \\
			GradDropout ($p$=0.3)
			& 87.84$_{\pm 0.41}$
			& 0.0671$_{\pm 0.0146}$ & 0.1705$_{\pm 0.0533}$
			& 0.1588$_{\pm 0.0495}$ & 0.1645$_{\pm 0.0678}$
			& 0.2204$_{\pm 0.0296}$ &0.1563 \\
			\midrule
			NODE ($\alpha$ = 0.2) & \underline{90.30}$_{\pm 0.14}$      &  {0.0256}$_{\pm 0.0096}$ &  {0.0305}$_{\pm 0.0169}$ &  0.0190$_{\pm 0.0048}$ &  \underline{0.2422}$_{\pm 0.0770}$ & 0.1611$_{\pm 0.1099}$ &\underline{0.0957}\\
			NODE ($\alpha$ = 0.1) & 89.31$_{\pm 0.02}$      &  0.0272$_{\pm 0.0086}$ &  \underline{0.0302}$_{\pm 0.0120}$ &  \underline{0.0183}$_{\pm 0.0058}$ &  \textbf{0.0790}$_{\pm 0.0265}$ & \underline{0.1561}$_{\pm 0.1080}$ &\textbf{0.0622} \\
			\bottomrule
	\end{tabular}}%
	\label{tab:cifar10_results}%
\end{table*}%

\begin{figure*}[thp]
	\centering
	\includegraphics[width=.75\textwidth]{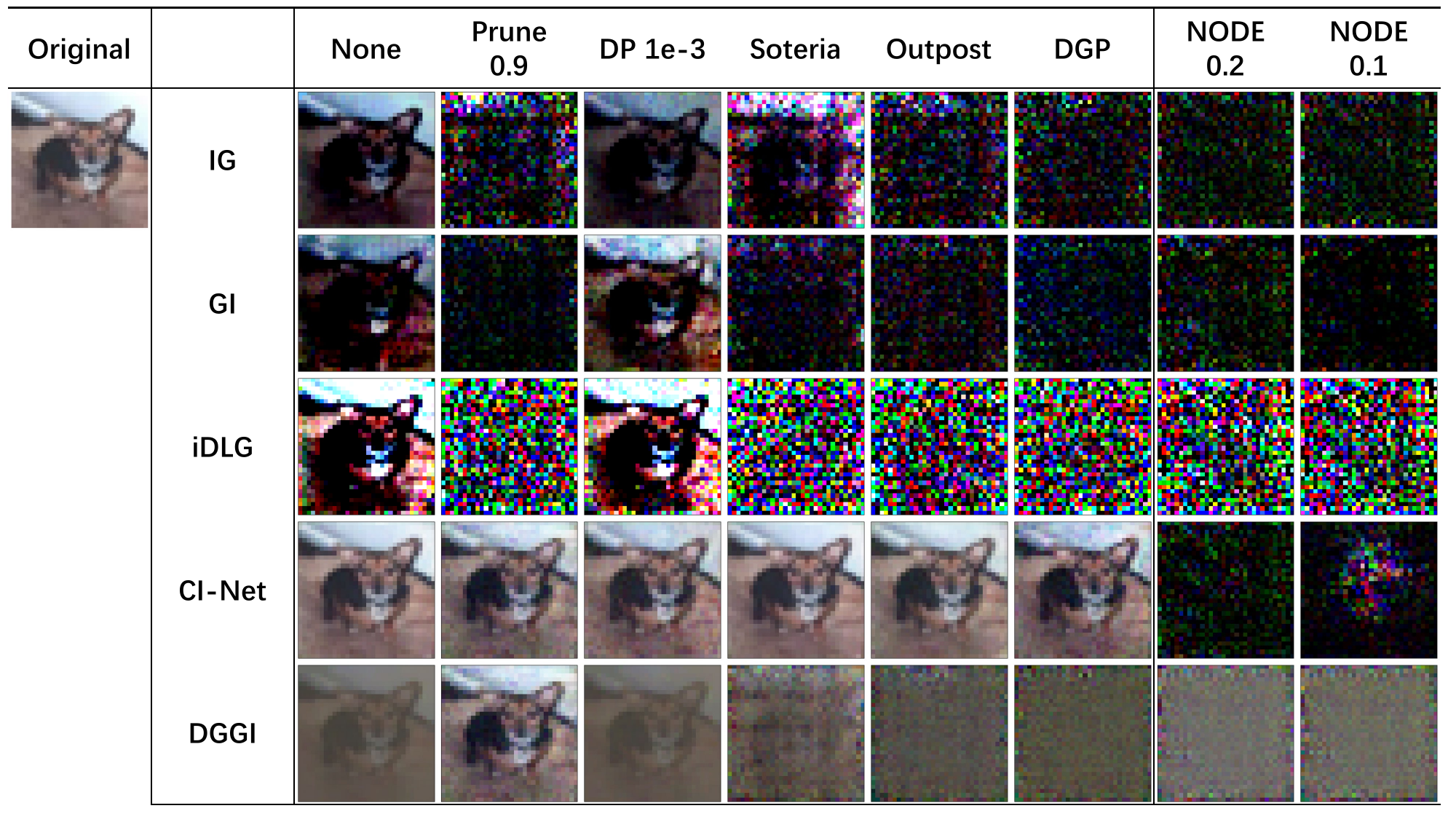}
	\caption{
		\centering
		Comparison between the original image and its reconstructions on CIFAR-10 across different combinations of attack and defense methods. Rows denote the attack applied, while columns illustrate the corresponding defense strategies.}
	\label{fig:recover_cifar10_untrain}
\end{figure*}

Figure~\ref{fig:recover_cifar10_untrain} shows a grid of recovered images under different defense methods for CIFAR-10. Each column corresponds to a specific defense method, and each row represents a particular inversion attack. The baseline reconstructions preserve substantial detail under the CI-Net attack, a generative approach capable of handling large batch sizes. Notably, in the absence of defense, it can fully recover the original training images, underscoring the data leakage risks posed by gradient-inversion attacks. In particular, DP with a noise scale of 1e-3 shows poor defense performance, whereas Outpost and DGP perform better. NODE’s reconstructions exhibit significant degradation, demonstrating its effectiveness in obstructing adversarial recovery while maintaining a high model accuracy.

\subsection{Chest X-ray Results}
We evaluate our methods on the Chest X-ray dataset~\cite{kermany_identifying_2018}; results are summarized in Table~\ref{tab: chest}. The unprotected baseline attains a test accuracy of $91.19\%$, but its predictions are prone to reconstruction via inversion attacks. Previous defense approaches reduce recoverability to some extent; however, they often introduce instability in training and substantially degrade predictive performance. In contrast, NODE with $\alpha=0.2$ strikes a favorable balance, yielding $89.79\%$ test accuracy while markedly reducing SSIM values across attack methods. Pushing $\alpha$ down to $0.1$ dramatically suppresses reconstruction quality: for CI-Net, the SSIM falls from $0.2857$ to $0.0579$, though stronger protection comes with an additional accuracy drop of $0.53\%$. Hence $\alpha=0.1$ is recommended for data-sensitive cases. 

\begin{table*}[h]
	\centering
	\caption{Performance on Chest X-ray: Final test accuracy and average SSIM under five gradient inversion attacks (GI, IG, iDLG, CI-Net and DGGI) across various defense methods. The best performance is in bold, and the second-best is underlined.}
	\resizebox{0.92\linewidth}{!}{
		\begin{tabular}{l|l|lllll}
			\hline
			\multicolumn{1}{c|}{\multirow{2}{*}{Defense   Methods}} & \multicolumn{1}{c|}{Performance} & \multicolumn{5}{c}{Attack (SSIM)}\\
			\cline{2-7}
			\multicolumn{1}{c|}{}       & \multicolumn{1}{c|}{Test Acc $\uparrow$}    & \multicolumn{1}{c}{GI $\downarrow$} & \multicolumn{1}{c}{IG $\downarrow$} & \multicolumn{1}{c}{iDLG $\downarrow$} & \multicolumn{1}{c}{CI-Net $\downarrow$} & \multicolumn{1}{c}{DGGI $\downarrow$} \\
			\hline
			None     & \textbf{91.19}$_{\pm 1.16}$ & 0.1727$_{\pm 0.1407}$  & 0.3332$_{\pm 0.0578}$  & 0.2041$_{\pm 0.0059}$    & 0.9505$_{\pm 0.0268}$      & 0.7574$_{\pm 0.1686}$    \\
			Pruning   & 86.11$_{\pm 0.60}$ & 0.0380$_{\pm 0.0334}$  & 0.0275$_{\pm 0.0069}$  & 0.0098$_{\pm 0.0023}$    & 0.3602$_{\pm 0.1014}$      & 0.1192$_{\pm 0.0184}$    \\
			DPSGD ($\sigma$=1e-3)   & 85.15$_{\pm 0.66}$ & 0.1611$_{\pm 0.0475}$  & 0.2961$_{\pm 0.0838}$  & 0.1587$_{\pm 0.0473}$    & 0.8022$_{\pm 0.1507}$      & 0.6201$_{\pm 0.1186}$    \\
			DPSGD ($\sigma$=4e-3)   & 84.99$_{\pm 0.66}$ & 0.0831$_{\pm 0.0320}$  & 0.0325$_{\pm 0.0079}$  & 0.0378$_{\pm 0.0117}$    & 0.1778$_{\pm 0.0670}$      & 0.2764$_{\pm 0.0662}$    \\
			Soteria  & 89.01$_{\pm 0.50}$   & 0.0203$_{\pm 0.0090}$  & 0.0369$_{\pm 0.0106}$  & 0.0148$_{\pm 0.0046}$    & 0.9624$_{\pm 0.0135}$      & 0.1827$_{\pm 0.0493}$    \\
			Outpost  & 87.90$_{\pm 0.88}$ & 0.0261$_{\pm 0.0137}$  & 0.0290$_{\pm 0.0057}$  & 0.0126$_{\pm 0.0041}$    & 0.9171$_{\pm 0.0633}$      & 0.1511$_{\pm 0.0462}$    \\
			DGP      & 89.18$_{\pm 0.88}$ & 0.0203$_{\pm 0.0151}$  & 0.0194$_{\pm 0.0025}$  & 0.0103$_{\pm 0.0035}$    & 0.8484$_{\pm 0.0731}$      & \underline{0.1107}$_{\pm 0.0527}$    \\
			GradientDropOut      & 86.24$_{\pm 0.44}$ & 0.0314$_{\pm 0.0181}$  & 0.0320$_{\pm 0.0014}$  & 0.0172$_{\pm 0.0028}$    & \underline{0.0784}$_{\pm 0.0345}$      & 0.1427$_{\pm 0.0568}$    \\
			\hline
			NODE ($\alpha$ = 0.2)     & \underline{89.79}$_{\pm 0.33}$ & \textbf{0.0143}$_{\pm 0.0045}$  & \underline{0.0131}$_{\pm 0.0037}$  & \underline{0.0095}$_{\pm 0.0027}$    & 0.2857$_{\pm 0.1184}$      & 0.1137$_{\pm 0.0593}$    \\
			NODE ($\alpha$ = 0.1)     & 89.26$_{\pm 0.39}$ & \underline{0.0144}$_{\pm 0.0054}$  & \textbf{0.0120}$_{\pm 0.0043}$  & \textbf{0.0085}$_{\pm 0.0035}$    & \textbf{0.0579}$_{\pm 0.0267}$      & \textbf{0.1087}$_{\pm 0.0594}$   \\
			\hline
	\end{tabular}
	}
	\label{tab: chest}
\end{table*}

\subsection{Visualization of Defense on X-Ray dataset}
\begin{figure}[htbp]
	\centering
	\includegraphics[width=0.6\textwidth]{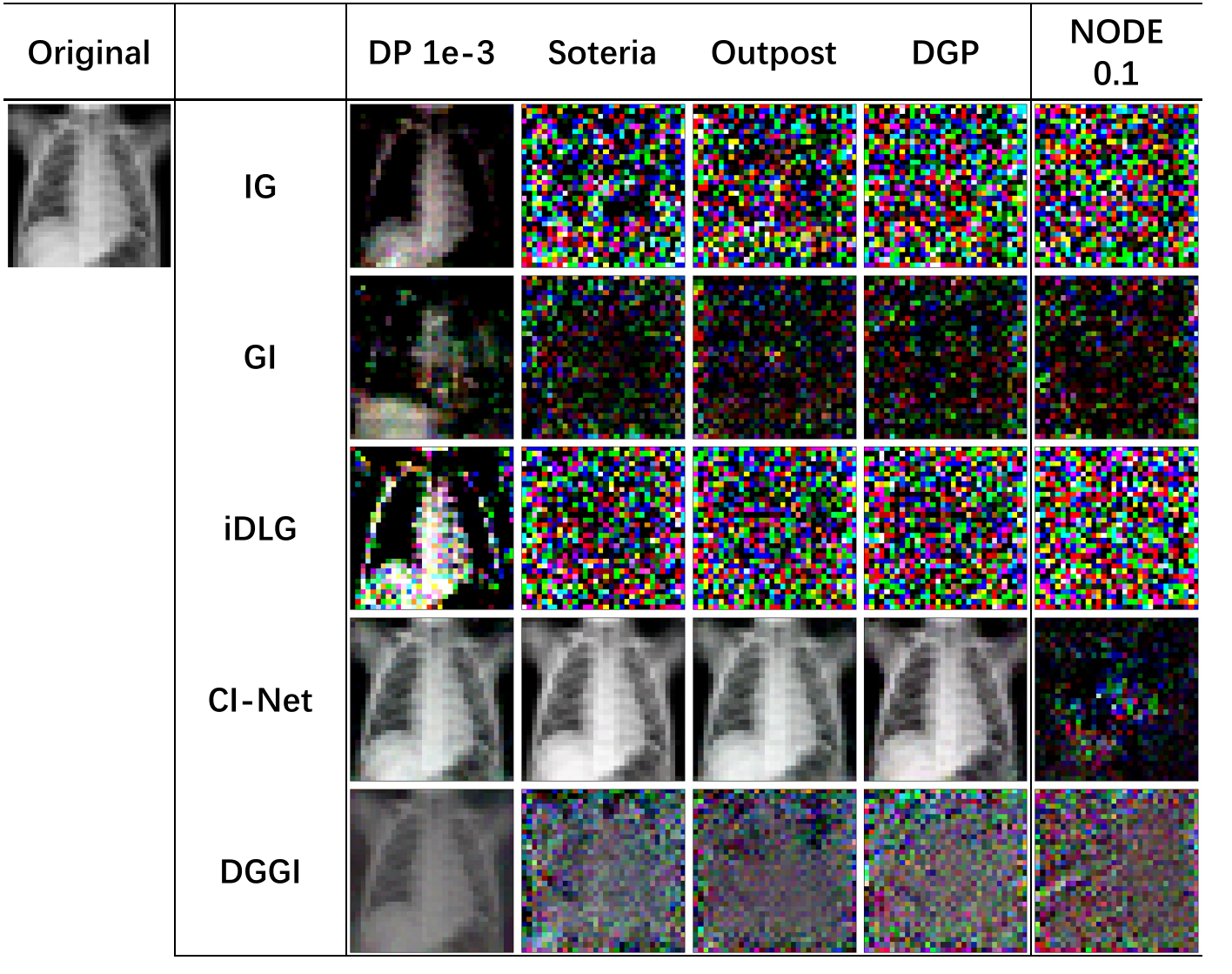}
	\caption{
		\centering
		Comparison of the original image and recovered images under different attack and (partial) defense methods for Chest X-ray.}
	\label{fig:recover_chest_untrain}
\end{figure}

In addition to the quantitative results measured by SSIM, Figure~\ref{fig:recover_chest_untrain} provides a qualitative comparison of various defense methods against five representative gradient inversion attacks. As shown in the figure, DPSGD fails to sufficiently obscure sensitive information, with several attacks recovering recognizable anatomical structures of the original chest X-ray image. Although Soteria, Outpost, and DGP effectively suppress reconstruction under IG, GI, and iDLG, they remain vulnerable to more advanced attacks, particularly CI-Net, which recovers the original anatomical structures with considerable fidelity. In contrast, NODE consistently disrupts image reconstruction across all five attacks, producing predominantly noise-like images with no clearly recognizable anatomical structures. The advantage of NODE is particularly pronounced against CI-Net, where all competing defense methods exhibit substantial visual information leakage. These qualitative observations further corroborate the quantitative SSIM results, demonstrating the effectiveness of NODE against gradient inversion attacks with diverse reconstruction mechanisms.

\subsection{ImageNet Defense Results}
In this section, we further present our experiments on the ImageNet dataset, but focus on the defense effectiveness (Table~\ref{tab:imagenet}). NODE attains low SSIM across attacks, indicating scalability to large-scale, high-resolution imaging scenarios. 
NODE’s reconstructions show its effectiveness in obstructing adversarial recovery throughout different attack methods while maintaining high model accuracy.
In this section, we further present FSIM values alongside additional recovered images to provide a more comprehensive understanding of the defense performance. We also discuss the training dynamics and time cost associated with each method to evaluate their efficiency. Additionally, we conduct ablation studies to examine the necessity of near-orthogonal projections, highlighting their importance in improving the robustness of the defense mechanisms against gradient inversion attacks.
\begin{table*}[h]
	\centering
	\caption{Performance on ImageNet: Average SSIM (with standard deviation) under five gradient inversion attacks (GI, IG, iDLG, CI-Net, and DGGI) across various defense methods. The best performance is in bold, and the second-best is underlined.}
    \resizebox{0.92\linewidth}{!}{
\begin{tabular}{l|ccccc|c}
	\toprule
	\multicolumn{1}{c|}{\multirow{2}{*}{Defense Methods}} & \multicolumn{6}{c}{Attack (SSIM)} \\
	\cline{2-7}          & \multicolumn{1}{c}{GI} & \multicolumn{1}{c}{IG} & \multicolumn{1}{c}{iDLG} & \multicolumn{1}{c}{CI-Net} & \multicolumn{1}{c}{DGGI} & Mean\\
	\midrule
	None  
	& 0.1087$_{\pm 0.0318}$ & 0.0963$_{\pm 0.0230}$ & 0.0388$_{\pm 0.0073}$ & 0.6531$_{\pm 0.0896}$ & 0.3897$_{\pm 0.0382}$ & 0.2573 \\
	
	Pruning (p=0.9) 
	& 0.0841$_{\pm 0.0315}$ & 0.0548$_{\pm 0.0203}$ & 0.0049$_{\pm 0.0003}$ & 0.1091$_{\pm 0.0154}$ & 0.0259$_{\pm 0.0259}$ & 0.0558 \\
	
	
	DPSGD ($\sigma$=1e-3)  
	& 0.1698$_{\pm 0.0404}$ & 0.0142$_{\pm 0.0018}$ & 0.0045$_{\pm 0.0004}$ & 0.1430$_{\pm 0.0489}$ & 0.1813$_{\pm 0.0202}$ & 0.1026 \\
	
	DPSGD ($\sigma$=4e-3)  
	& 0.1157$_{\pm 0.0390}$ & 0.0054$_{\pm 0.0004}$ & 0.0048$_{\pm 0.0002}$ & 0.0336$_{\pm 0.0064}$ & \underline{0.0196}$_{\pm 0.0010}$ & 0.0358 \\
	
	Soteria 
	& 0.1017$_{\pm 0.0350}$ & 0.0091$_{\pm 0.0010}$ & 0.0047$_{\pm 0.0003}$ & 0.0705$_{\pm 0.0137}$ & 0.0713$_{\pm 0.0067}$ & 0.0516 \\
	
	Outpost 
	& 0.0877$_{\pm 0.0408}$ & 0.0564$_{\pm 0.0194}$ & 0.0060$_{\pm 0.0008}$ & 0.1715$_{\pm 0.0300}$ & 0.3388$_{\pm 0.0238}$ & 0.1321 \\
	
	DGP   
	& 0.1012$_{\pm 0.0158}$ & \textbf{0.0054}$_{\pm 0.0004}$ & 0.0046$_{\pm 0.0002}$ & 0.0525$_{\pm 0.0181}$ & \textbf{0.0106}$_{\pm 0.0014}$ & 0.0349 \\
	
	\midrule
	NODE ($\alpha=0.2$) 
	& \underline{0.0652}$_{\pm 0.0298}$ & 0.0072$_{\pm 0.0010}$ & \underline{0.0044}$_{\pm 0.0002}$ & \underline{0.0208}$_{\pm 0.0037}$ & 0.0578$_{\pm 0.0206}$ & \underline{0.0311} \\
	
	NODE ($\alpha=0.1$) 
	& \textbf{0.0572}$_{\pm 0.0230}$ & \underline{0.0058}$_{\pm 0.0004}$ & 0.0046$_{\pm 0.0002}$ & \textbf{0.0167}$_{\pm 0.0023}$ & 0.0320$_{\pm 0.0096}$ & \textbf{0.0233} \\
	
	\bottomrule
\end{tabular}}
	\label{tab:imagenet}%
\end{table*}%



\subsection{Ablation Study: Impact of Cosine Similarity}
Our proposed method projects the original gradients onto a nearly orthogonal direction by injecting noise perpendicular to the original gradients. To demonstrate the necessity of this near-orthogonal design, we conduct the following ablation studies.

\begin{table}[h]
	\centering
	\caption{Test accuracy and average SSIM value under CI-Net attack for NODE, with different cosine similarity $\alpha$ on CIFAR-10.}
	\begin{tabular}{l|c|c|c|c}
		\toprule
		Method & Angle $\theta$ & SNR (dB) & Test Acc $\uparrow$ & CI-Net $\downarrow$ \\
		\midrule
		$\alpha$ = 0.7 & $45.57^\circ$ &-0.0869 & 91.23$_{\pm 0.14}$ & 0.9570$_{\pm 0.0231}$ \\
		$\alpha$ = 0.3 & $72.54^\circ$ &-5.0240 & 90.74$_{\pm 0.24}$ & 0.5610$_{\pm 0.1163}$ \\
		$\alpha$ =  0.2 & $78.46^\circ$ &-6.9011 & 90.30$_{\pm 0.14}$ & 0.2422$_{\pm 0.0770}$ \\
		$\alpha$ = 0.1 & $84.26^\circ$ &-9.9782  & 89.31$_{\pm 0.02}$ & 0.0790$_{\pm 0.0265}$ \\
		\bottomrule
	\end{tabular}%
	\label{tab:ablation}%
\end{table}%

We vary the noise scale to manipulate the cosine similarity $\alpha$ between the defended gradient $g'$ and the original gradient $g$. As shown in Table~\ref{tab:ablation}, a high cosine similarity (i.e., a small angle) corresponds to less aggressive noise injection, which results in higher SSIM values but weaker defense against inversion attacks. For instance, the CI-Net can nearly recover the original data with an SSIM value of $0.9570$, which violates the safety principles outlined in Theorem~\ref{thm:defense}, as this leads to a smaller lower bound for $\ell_1$ (or $\sigma$). In contrast, a larger angle (e.g., $84.26^\circ$ with $\alpha = 0.1$) strengthens the defense against gradient inversion attacks, reducing the CI-Net’s SSIM to 0.0790. While this offers better protection, it comes at the expense of a slight decrease in model performance. But in general, setting $\alpha$ to 0.1 is recommended for most applications, especially for the privacy-sensitive ones.

\section{Conclusion}
We proposed a unified controlled-dynamics formulation that subsumes a broad class of gradient-modification defenses and permits unified convergence and safety analyses, removing the need for method-specific proofs. Within this framework, we introduced ``Near Orthogonal graDient projEction'' (NODE), which injects noise solely in the subspace orthogonal to the true gradient to maintain a fixed cosine similarity between the defended and true gradients. Under standard assumptions and fixed learning rates, NODE preserves the first-order descent direction and provably maintains convergence. Empirically, NODE significantly reduces the success of diverse gradient inversion attacks on different datasets while retaining high task performance, thus delivering a practical utility–privacy trade-off for privacy-sensitive applications.

\clearpage
\bibliographystyle{unsrtnat}
\bibliography{references}  


\appendix
\onecolumn

\section{Proof for Theorems}
\subsection{Proof for Theorem~\ref{thm: convergence} and Theorem~\ref{thm: convergence nonconvex}}
Theorem~\ref{thm: convergence} and~\ref{thm: convergence nonconvex} establish a convergence result under standard convexity assumptions, with its proof derived from foundational convex analysis principles. For a detailed treatment of the relevant concepts, the proof can be found in a series of classical works on convex analysis, such as Sections 4.2 and 4.3 of~\cite{bottou2018optimization}, which provides a comprehensive overview of the techniques used.
\subsection{Proof for Remark~\ref{remark: perpendicular}}
\paragraph{Proof} From convexity of loss $L$, we have:
\begin{equation}
	L(y)\geq L(x)+\nabla L(x)^T(y-x),\forall x,y\in\R^n
\end{equation}
If the defended gradient is perpendicular to the true gradient, i.e., $\nabla L(x)^Tg=0$, we can let $y=x-\eta g$ and get the desired result:
\begin{equation}
	L(x-\eta g)\geq L(x).
\end{equation}
\subsection{Proof for Theorem~\ref{thm: in-grad Lipschitz}}
\paragraph{Proof} From Lipschitz continuity, we have 
\begin{equation}
	\|G_w(x)-G_w(x')\|_2\leq l_g\|x-x'\|_2.
\end{equation}
Substitute $G_w(x')$ with $g'=C(g,x)$ and $x'$ with an element in $ G_w^{-1}(C(g,x))$, we have:
\begin{equation}
	\frac{1}{l_g}\|g-g'\|_2\leq \|x-x'\|_2,\quad \forall x'\in G_w^{-1}(g')
\end{equation}
thus
\begin{equation}
	\frac{1}{l_g}\|g-g'\|_2\leq\min_{x'\in G_w^{-1}(g')}\|x-x'\|_2,
\end{equation}

Taking the minimum of $x\in D$ in both sides, we will get:
\begin{equation}
	\min_{x\in D}\frac{1}{l_g}\|g-C(g,x)\|_2\leq\min_{x\in D}\min_{x'\in G_w^{-1}(g')}\|x-x'\|_2=\ell_1 
\end{equation}
\subsection{Proof for Theorem~\ref{thm: DP converge}}
\paragraph{Proof} Since the noise added by DPSGD is zero-mean, we have $\E[g']=\nabla L(w)$, thus:
\begin{align}
	\nabla L(w)^\top\E[g']=\|\nabla L(w)\|_2^2,
\end{align}
i.e. $\mu=1,\mu_g=1$.
Further, the variance of the defended gradient $g'$ is the variance of the noise, thus:
\begin{equation}
	\text{Var}(g')=\text{Var}(\xi)=a^2\sigma^2,
\end{equation}
which means $M=a^2\sigma^2,M_V=0$
Substitute $\mu,M$ into Eq. \eqref{eq: converge} we can get the result.
\subsection{Proof for Theorem~\ref{thm:ell_1 for DPSGD}}
\paragraph{Proof} Denote $\xi_t=a\sigma X$, where $X\sim N(0,I_m)$.
From Theorem 3.1.1 in ~\cite{Vershynin2018HighDimensionalP}, we obtain
\begin{equation}
	\mathbb{P}\{|\|X\|_2-\sqrt{m}|\geq t \}\leq 2\exp(-\frac{kt^2}{K^4}),
\end{equation}
where $K$ is the $\psi_2$ norm of $X_i\sim N(0,1) $ i.i.d, which is a constant. $k$ is another absolute constant.
Substitute $\exp(-\frac{kt^2}{K^4})$ with $\delta$, we know 
\begin{equation}
	\|X\|_2 \geq \sqrt{m}-\frac{K^2}{\sqrt{k}}\sqrt{\log \delta^{-1}}
\end{equation}
holds with probability of $1-2\delta$, i.e. 
\begin{equation}
	\|C(g,x)-g\|_2=a\sigma\|X\|_2\geq a\sigma\left(\sqrt{m}-a_0\sqrt{\log \delta^{-1}}\right)
\end{equation}
with probability of $1-2\delta$, where $a_0=\frac{K^2}{\sqrt{k}}$ is a constant. Then the proof is finished with Theorem \ref{thm: in-grad Lipschitz}.

\subsection{Proof for Theorem~\ref{thm:ell_0 for Pruning}}
\paragraph{Proof}
Since we always prune the small value and there is no randomness in the pruning method, we can get:
\begin{align}
	\nabla L(w)^\top\E[g']&\geq p\|\nabla L(w)\|_2^2,\\
	\|E[g']\|_2&\leq\|\nabla L(w)\|_2,\\
	Var(g')&=0,
\end{align}
where $p$ is the pruning rate. Then we can get $\mu\geq p>0,\mu<\mu_G\leq 1,M=0=M_G$, and thus $\ell_{0,\text{Pruning}}=0$. Even if we consider the random prune, we have $Var(g')\leq\|g'\|_2\leq\|g\|_2^2$, and other coefficients remain the same, so we can still derive the same result. 
\subsection{Proof for Theorem~\ref{thm:ell_1 for Pruning}}
\paragraph{Proof} Similar to Proof for Theorem~\ref{thm:ell_1 for DPSGD}, we denote $g=\sigma X$, where $X\sim N(0,I_m)$.
Use Theorem 3.1.1 in ~\cite{Vershynin2018HighDimensionalP} for $\hat{X}$, we obtain
\begin{equation}
	\mathbb{P}\left\{\left|\frac{\|\hat{X}\|_2}{\sqrt{V(a)}}-\sqrt{m}\right|\geq t \right\}\leq 2\exp\left(-\frac{kt^2}{K^4}\right),
\end{equation}
where $K$ and $k$ is the same constants as in the proof of Theorem \ref{thm:ell_1 for DPSGD}. 
Let $s=t\sqrt{V(a)},\delta=\exp(-\frac{kt^2}{K^4})$, and use Theorem \ref{thm: in-grad Lipschitz}, we finish the proof in the same way.

\subsection{Proof for Theorem~\ref{thm:ell_0 for NODE}}
\paragraph{Proof}
Similar as the proof of the Thoerem~\ref{thm:ell_0 for NODE}, we need to analyze the parameters in Eq. \eqref{eq: converge}. Since the noise is uniformly distributed, its expectation is 0, we have.
\begin{equation}
	\nabla L(w)^\top\E[g']=\|\nabla L(w)\|_2^2,\quad \|\E[g']\|_2=\|\nabla L(w)\|_2
\end{equation}
i.e. $\mu=1=\mu_G$.
Further, the angle between defended gradient $g'$ and $g$ is fixed, so the norm of defended gradient $g'$ is fixed, i.e.
\begin{equation}
	\|g'\|_2\equiv\sec\theta\|g\|_2.
\end{equation}
Therefore, 
\begin{equation}
	Var(g')=\E[\|g'\|_2^2]-\|\E[g']\|_2^2=\tan^2\theta\|g\|_2^2,
\end{equation}
which means $M=0,M_V=\tan^2\theta$. From Theorem \ref{thm: convergence} and Eq. \ref{eq: converge}, we have $\ell_{0,\text{NODE}}=0$. 

\subsection{Proof for Theorem~\ref{thm: lyapunov}}
\paragraph{Proof} 
A Lyapunov function is a scalar function \( V(w):\R^m\to\R \) used to study the stability of dynamical systems. Given a dynamical system of the form

\[
\frac{dw}{dt} = f(w),
\]
where \( w \) represents the state of the system and \( f(w) \) is the system's evolution over time, a Lyapunov function \( V(w) \) satisfies the following conditions:

\begin{enumerate}
	\item Positive Definite: \( V(w) \geq 0 \) for all \( w \), and \( V(w) = 0 \) only at the equilibrium point (usually \( w = 0 \)).
	\item Decreasing along system trajectories: The time derivative of \( V(w) \), denoted as
	
	\[
	\frac{d V(w)}{dt} = \nabla V(w) \cdot f(w),
	\]
	
	should be negative definite or negative semi-definite. This condition implies that the Lyapunov function decreases over time, indicating that the system is ``moving towards'' the equilibrium point.
\end{enumerate}

If these conditions are satisfied, the equilibrium point is stable. Specifically:

\begin{itemize}
	\item Asymptotic Stability: Occurs if \( \frac{d V(w)}{dt} \) is negative definite, meaning that \( V(w) \) strictly decreases over time.
	\item Stability in the sense of Lyapunov: Holds if \( \frac{d V(w)}{dt} \) is negative semi-definite, meaning that \( V(w) \) either decreases or remains constant.
\end{itemize}

For the training dynamics Eq. \eqref{eq: modified dynamics}, it is easy to prove that the equilibrium points are the saddle points of the loss function $L(w)$, which is also the minima in the strongly convex case. Therefore, the convergence to optimal loss is equivalent to the stability property in the control theory.

To prove $V:=(L(w)-L^*)^2$ is the Lyapunov function for the NODE method, we need to verify that (1) $V(w^*)=0$; (2) $V(w)>0,\forall w\neq w^*$; (3) $\frac{dV}{dt}<0$. 

For strongly-convex $L(w)$, it is easy to verify that the first two conditions are satisfied. We only need to prove the third condition:
\begin{equation}
	\begin{aligned}
		\frac{dV}{dt} &= 2 (L(w(t))-L^*)\frac{dL}{dt}\\
		&= 2 (L(w(t))-L^*)\nabla L(w(t))^\top\frac{dw}{dt}\\
		&= -2 (L(w(t))-L^*)\|\nabla L(w(t))\|^2_2<0.
	\end{aligned}
\end{equation}
Consequently, $(L(w)-L^*)^2$ is a Lyapunov function for the NODE method, and it will asymptotically reach minimum $0$, as claimed in the theorem.

\subsection{Proof for Theorem~\ref{thm:defense}}
\textbf{Proof:} Under Assumption \ref{assumption:gradient norm}, the gradient norm threshold can be taken as $a= b\|\nabla L(w)\|_2$ and the noise scale $a^2\sigma^2=\tan^2\theta\|\nabla L(w)\|_2^2$. Therefore $\sigma=\frac{\tan\theta\|\nabla L(w)\|_2}{a}\geq \frac{\tan\theta}{b}$.

\section{Additional Experiments}
The part contains additional experiments to support Section~\ref{sec:5}.

\subsection{Detailed Experimental Setup}
We evaluate the proposed method in a simulated federated learning environment. Following prior studies on gradient inversion attacks, we adopt ResNet-18 models with sigmoid activations, which have been shown to be more vulnerable to gradient leakage compared with commonly used activation functions. The federated setting consists of three simulated clients, where each client performs one local optimization step before communicating gradients with the server. Models are trained using stochastic gradient descent (SGD) without momentum, with a batch size of 16. The initial learning rate is set to 0.1 and gradually reduced according to a cosine decay schedule throughout training.

\subsection{FSIM Results of the Attacks}
In addition to SSIM, we further evaluate the defense methods using the Feature Similarity Index (FSIM) to assess the reconstructed image quality. 

\begin{table}[h]
	\centering
	\caption{Average FSIM on CIFAR10, Tiny-ImageNet, Chest X-ray, and ImageNet under the CI-Net attack. The best is in bold, and the second is underlined.}
	\resizebox{0.8\linewidth}{!}{
		\begin{tabular}{l|cccc|c}
			\toprule
			Defense Methods & CIFAR-10 & Tiny-ImageNet & Chest X-ray & ImageNet & Mean\\
			\midrule
			None  
			& 0.9843$_{\pm 0.0083}$ & 0.9899$_{\pm 0.0032}$ & 0.6725$_{\pm 0.1704}$ & 0.6483$_{\pm 0.1217}$ & 0.8238\\
			
			Pruning (p=0.9) 
			& 0.9117$_{\pm 0.0257}$ & 0.9313$_{\pm 0.0146}$ & 0.6225$_{\pm 0.1364}$ & 0.5732$_{\pm 0.0817}$ & 0.7597\\
			
			Pruning (p=0.99) 
			& 0.6602$_{\pm 0.0363}$ & \underline{0.7012}$_{\pm 0.0417}$ & 0.6429$_{\pm 0.0153}$ & 0.5721$_{\pm 0.0838}$ & 0.6441\\
			
			DP ($\sigma$=1e-3) 
			& 0.9394$_{\pm 0.0256}$ & 0.9607$_{\pm 0.0062}$ & 0.6452$_{\pm 0.1321}$ & 0.5589$_{\pm 0.0900}$ & 0.7761\\
			
			DP ($\sigma$=4e-3) 
			& 0.6796$_{\pm 0.0451}$ & 0.7486$_{\pm 0.0188}$ & 0.5569$_{\pm 0.0425}$ & 0.5102$_{\pm 0.0814}$ & 0.6238\\
			
			Soteria 
			& 0.9864$_{\pm 0.0064}$ & 0.9901$_{\pm 0.0024}$ & 0.6408$_{\pm 0.1680}$ & 0.6408$_{\pm 0.1680}$ & 0.8145\\
			
			Outpost 
			& 0.6353$_{\pm 0.1613}$ & 0.7244$_{\pm 0.1297}$ & 0.6296$_{\pm 0.1493}$ & 0.5783$_{\pm 0.0836}$ & 0.6419\\
			
			DGP 
			& 0.6284$_{\pm 0.1324}$ & 0.6490$_{\pm 0.1472}$ & 0.6176$_{\pm 0.1350}$ & \textbf{0.4993}$_{\pm 0.0787}$ & \underline{0.5986}\\
			
			\midrule
			NODE ($\alpha$=0.2) 
			& \underline{0.6521}$_{\pm 0.0479}$ & 0.7507$_{\pm 0.0426}$ & \underline{0.5803}$_{\pm 0.0533}$ & 0.5236$_{\pm 0.0795}$ & 0.6267\\
			
			NODE ($\alpha$=0.1) 
			& \textbf{0.5731}$_{\pm 0.0342}$ & \textbf{0.5993}$_{\pm 0.0240}$ & \textbf{0.5546}$_{\pm 0.0279}$ & \underline{0.5081}$_{\pm 0.0721}$ & \textbf{0.5588}\\
			
			\bottomrule
		\end{tabular}
}%
	\label{tab:fsim}%
\end{table}%

Table~\ref{tab:fsim} reports the FSIM values for different defense algorithms under the CI-Net attack. In the CIFAR-10 dataset, the baseline performance without any defense yields a high FSIM value of 0.9843, suggesting that the model is vulnerable to data leakage through gradient inversion attacks. Such high values indicate that without proper defense mechanisms, the model's gradients retain too much information, allowing for potential reconstruction of original data. Among the defense methods tested, the NODE technique stands out as the most effective, achieving the lowest FSIM value of 0.5731, signaling strong protection against inversion attacks. It outperforms other methods, including Soteria, which follows closely with a high FSIM value of 0.9864. Therefore, NODE demonstrates the best defense on CIFAR-10, significantly reducing the risk of data leakage while maintaining robust defense performance.

In the Tiny-ImageNet dataset, NODE with $\alpha=0.1$ continues to show significant improvement in defense against gradient inversion attacks, achieving the best FSIM value of 0.5993. This indicates a strong reduction in the ability to reconstruct the original data, highlighting the effectiveness of the method. However, methods like pruning with a large ratio ($p=0.99$) and differential privacy (DP) with higher noise levels ($\sigma$=4e-3) can potentially outperform NODE with $\alpha=0.2$ in terms of defense strength, with FSIM values of 0.7012 and 0.7486, respectively. The defense performance of NODE with $\alpha=0.2$ is slightly higher than these two methods, but yields the best test accuracy $53.27\%$ in this scenario.

%

\subsection{Efficiency Analysis}
\begin{figure}[h]
	\centering
	\includegraphics[width=0.45\linewidth]{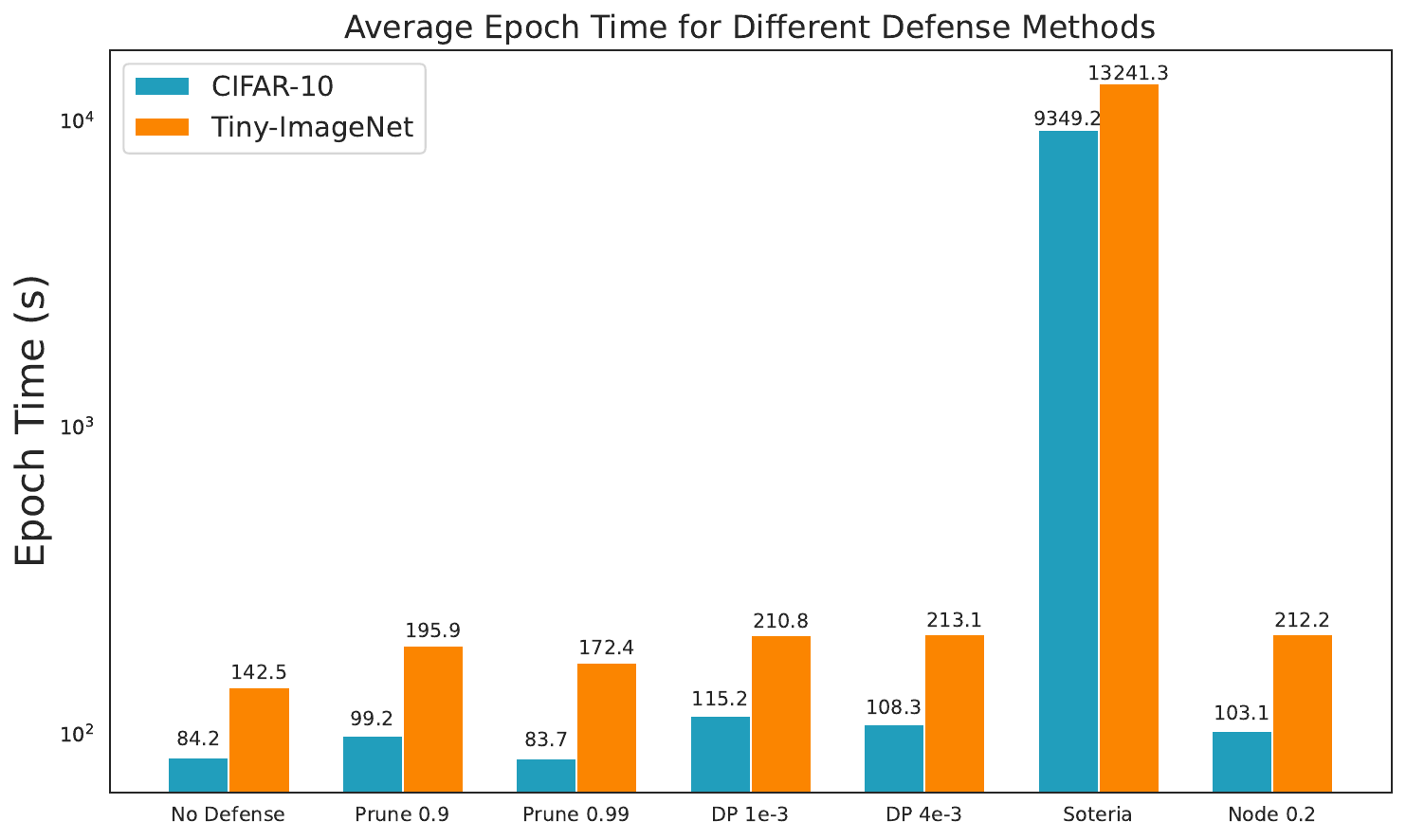}
	\caption{
		\centering
		Average running time for one epoch of each defense method.}
	\label{fig:efficiency_curve}
\end{figure}

In addition to accuracy and defense effectiveness, we assess the efficiency of our method in terms of training time. Figure~\ref{fig:efficiency_curve} compares the training duration per epoch across different defense methods. 

On the CIFAR-10 Dataset, we have the following conclusions:
\begin{itemize}
	\item The baseline method (No Defense) takes the least time (\(84.2\) seconds), making it almost the most efficient algorithm.
	\item Both Pruning (p=$0.9$) (\(99.2\) seconds) and Pruning (p=$0.99$) (\(83.7\) seconds) introduce a modest increase in time. While pruning itself may introduce extra computational time, backpropagation with the pruned gradients generally leads to less time. As such, pruning does not lead to significant increases in defending gradient inversion.
	\item Differential privacy methods DP ($\sigma$=1e-3) and DP ($\sigma$=4e-3) have a noticeable increase in time, with values of \(115.2\) and \(108.3\) seconds, respectively.
	\item Soteria incurs a significant computational overhead with a time of \(9349.2\) seconds per epoch.
	\item Unlike Soteria, the Node method exhibits a moderate increase in time, with an average epoch time
	of $103.1$ seconds. This makes Node's time cost comparable to that of other methods, such as DP and pruning.
\end{itemize}

The results on the Tiny-ImageNet dataset are consistent with the above conclusions. In general, our results indicate that NODE incurs equivalent or slightly less overhead than the DPSGD methods. This efficiency makes NODE particularly suitable for real-world federated learning applications where both performance and speed are critical.

\end{document}

%% file: math.tex
\newtheorem{theorem}{Theorem}[section]

\newtheorem{assumption}[theorem]{Assumption}
\newtheorem{remark}[theorem]{Remark}

\def\*#1{\boldsymbol{#1}}

\newcommand{\R}{\mathbb{R}}

\newcommand{\E}{\mathbb{E}}

